\documentclass{article}

 \usepackage[final, nonatbib]{neurips_2020}

\usepackage[utf8]{inputenc} 
\usepackage[T1]{fontenc}    
\usepackage{hyperref}       
\usepackage{url}            
\usepackage{booktabs}       
\usepackage{amsfonts}       
\usepackage{amsmath}        
\usepackage{mathtools}
\usepackage{subcaption}
\usepackage{nicefrac}       
\usepackage{microtype}      
\usepackage{graphicx}
\usepackage{todonotes}
\usepackage{macros}
\usepackage{booktabs}
\usepackage{algorithm}
\usepackage{algorithmicx}
\usepackage{algpseudocode}
\usepackage{diffcoeff}

\usepackage{subcaption}

\newcommand{\BlackBox}{\rule{1.5ex}{1.5ex}}  

\title{GENNI: Visualising the Geometry of Equivalences for Neural Network Identifiability}

\author{
 Daniel Lengyel \thanks{Equal contribution} \hspace{0.5pt} \\
 Imperial College London\\
 \texttt{dl2119@ic.ac.uk} \\
 \And 
 Janith Petangoda \footnotemark[1] \hspace{0.5pt} \\
 Imperial College London\\
 \texttt{jcp17@ic.ac.uk}\\
 \And 
 Isak Falk \footnotemark[1] \hspace{0.5pt}\\
 University College London\\
 \texttt{ucabitf@ucl.ac.uk} \\
 \And 
 Kate Highnam \footnotemark[1] \hspace{0.5pt}\\
 Imperial College London\\
 \texttt{kwh19@ic.ac.uk}
 \And 
 Michalis Lazarou \\
 Imperial College London\\
\texttt{ml6414@imperial.ac.uk}\\
 \And 
 Arinbjörn Kolbeinsson \\
 Imperial College London\\
 \texttt{ak711@ic.ac.uk}\\
 \And
 Marc Peter Deisenroth \\
 University College London\\
 \texttt{m.deisenroth@ucl.ac.uk}
 \And 
 Nicholas R. Jennings \\
 Imperial College London\\
\texttt{n.jennings@imperial.ac.uk}\\
}

\begin{document}

\maketitle

\begin{abstract}
We propose an efficient algorithm to visualise symmetries in neural networks. Typically, models are defined with respect to a parameter space, where non-equal parameters can produce the same input-output map. Our proposed method, GENNI, allows us to efficiently identify parameters that are functionally equivalent and then visualise the subspace of the resulting equivalence class. By doing so, we are now able to better explore questions surrounding identifiability, with applications to optimisation and generalizability, for commonly used or newly developed neural network architectures. 
\end{abstract}



\section{Introduction}

Confusion around identifiability\footnote{
This is not exactly the same as statistical identifiability, which traditionally refers to problems of inference \cite{casella2002statistical}. We are extending this term to discuss model equivalences in deep networks.
} and its consequences in machine learning, especially in deep learning, continues to critically impact aspects of model development. 
Intuitively, identifiability describes the situation where two models defined with distinct parameters in the model parameter space are functionally equivalent. Thus, when learning on the parameter space, one cannot uniquely identify a suitable solution for their problem\footnote{
We note that functionally distinct networks can be unidentifiable due to weaknesses in the loss function. In the present work, only the impact of the network architecture on identifiability is considered.
}.
Such cases can mislead design choices in inference \cite{pourzanjani2017improving} or optimisation \cite{amari1998natural}, and our understanding of generalisability of machine learning models \cite{Dinh2017, sagun2017empirical}. 

For deep learning in particular, the lack of identifiability in the parameter space can lead to multiple global minima \cite{Albertini1993Form, Albertini1993}. For example, this can arise from permutation symmetries, where swapping the incoming and outgoing weights across different pairs of neurons results in the same input-output map. Work in the literature handle these symmetries by a choice of a reparameterisation \cite{pourzanjani2017improving}, regularisation \cite{Draxler2018}, or natural gradients \cite{amari1998natural} to produce an optimisation scheme that is invariant under such symmetries.
However, this is only possible with an understanding of the transformation groups which generate such symmetries.
Often, identifying such symmetries is a tedious undertaking that involves trial and error, and an unreasonable analytical understanding of modern neural network architectures. 
While generally visualisations would be the first method by which to explore poorly understood structures, such intuition aiding tools have not been available. 

Through our tool, GENNI\footnote{The code is available on Github at \url{https://github.com/Do-Not-Circulate/GENNI}.}, we propose to visually and empirically find and explore symmetries of neural networks to guide analytical studies.
GENNI uses an optimisation-based search algorithm to both efficiently and systematically find subsets of high-dimensional parameter spaces that produce equivalent neural networks.
Plots of these parameters then immediately give insight into topological and geometric properties of these equivalent sets. 
We demonstrate how one could reason about identifiability using our visualisations by considering a simple fully connected network for which the symmetries are well understand. 
We believe that using GENNI, such analysis can then be applied to more complex models. 
Our work is in contrast to other methods which visualize the loss landscape \cite{li2018visualizing, Draxler2018}; GENNI is independent of the loss function chosen and finds only the symmetries inherent to the network architecture.


\section{Theory} \label{sec:theory}

Given a domain $\X$ and co-domain $\Y$, we want to find the set of parameters in parameter space $\Theta$ which are functionally equivalent under a model architecture $\phi : \X \times \Theta \rightarrow \Y$.
An equivalence class $[\theta]$ for a parameter $\theta \in \Theta$ is given by the equivalence relation $\theta_1 \sim \theta_2$ iff $\phi(\cdot, \theta_1) = \phi(\cdot, \theta_2)$.
To practically check whether $\theta_1 \sim \theta_2$, a metric on the space of functions (for which the parameters are fixed) is introduced: $d(\phi(\cdot, \theta_1), \phi(\cdot, \theta_2)) = \sqrt{ \int_{\X} \vert \phi(x, \theta_1) - \phi(x, \theta_2) \vert_2^2 \dl x}$.
Then due to the definition of a metric: $\phi(\cdot, \theta_1) = \phi(\cdot, \theta_2)$ iff $d(\phi(\cdot, \theta_1), \phi(\cdot, \theta_2)) = 0$; thus the equivalence holds. 

It is common to express equivalence classes in terms of a group of transformations that result in them; let us denote the group that corresponds to $[\theta]$ by $\G_\theta$. This group can be composed of several smaller groups that each represent a class of symmetries (e.g. the group of permutations $\G_{\mathrm{perm}}$). Then, given a set of groups $\{\G_i\}_{i \in \mathcal{I}}$ acting on a specific $\theta$, the equivalence class $[\theta]$ is the orbit \cite[Chapter~7]{lee2013smooth} of the action of the direct product group $G_1 \times ... \times G_i$, $\forall i \in \mathcal{I}$, denoted as $\G_{\theta}$ \footnote{The group operator is the canonical operator on a direct product of groups.  The group action here is given by the composition of transformations. That is, for $(g_1, ..., g_i) \in \G_\theta$, the action $\pi(\theta, (g_1, ..., g_i)) = g_1 \circ ... \circ g_i(\theta)$}. In order to form  $\G_\theta$, each $G_i$ must be identified. This can be difficult to do for modern neural networks, as they can be overtly over-parameterised with many complex layers (such as CNNs \cite{krizhevsky2012imagenet}, ResNets \cite{he2016deep}, transformers \cite{devlin2018bert}, etc.). 
Furthermore, the set of groups $\G_i$ is not equal for all $\theta$; \cite{petzka2020notes} showed this for a simple feedforward network. Finally, the structure of the orbit can be non-trivial, even once $\G_\theta$ is known. This could, for example, be due to non-trivial isotropy subgroups\footnote{An isotropy, or stabilizer subgroup $(\mathcal{S}_\G(\theta) \subseteq \G) = \{g \in \G | g(\theta) = \theta \}$, where $\theta \in \Theta$, and $g(\theta)$ is the action of $g$ on $\theta$.  It is trivial if $\mathcal{S}_\G(\theta) =\{e\}$, where $e$ is the neutral element of $\G$.} \cite[Chapter~7]{lee2013smooth} of the group action. 

Thus, the equivalence class of $\theta$, and by extension the quotient set $\Theta /_\sim$, can be difficult to study analytically. 
A visualisation of a single equivalence class could help derive the symmetries of various network architectures by looking at its topological and geometric properties, such as connectedness, dimension and curvature, among others. 
Following this, we can gain an understanding of the quotient set by looking at how different equivalence classes in a neighbourhood stack together. While this is useful \cite{meng2018mathcal, Draxler2018}, we presently focus on visualising a single equivalence class; extensions to understanding the quotient set is beyond the scope of this work.

\section{GENNI: Visualising the \textit{G}eometry of \textit{E}quivalences for \textit{N}eural \textit{N}etwork \textit{I}dentifiability}\label{sec:method}

GENNI\footnotemark[4] is our visualisation tool for studying complex symmetries of neural networks.

Since neural network parameter spaces are high-dimensional, it is difficult to find $[\theta]$ via brute-force methods such as exhaustive grid search or random guessing, as they scale exponentially with the number of dimensions. 
At a high level, GENNI uses stochastic gradient descent (SGD) to efficiently search for likely candidates of $[\theta]$, even as the dimension of $\Theta$ increases. 
More concretely, to find parameters which are in $[\theta]$ we randomly initialize vectors in $\Theta$ and use SGD to minimize the auxiliary loss $J(\tilde{\theta}) = \frac{1}{\abs{C}}\sum_{x \in C} \vert \phi(x, \theta) - \phi(x, \tilde{\theta})\vert_2^2$.
We use $J(\tilde{\theta})$ as an approximation of $d(\phi(\cdot, \theta), \phi(\cdot, \tilde{\theta}))$, as seen in the pseudo-code found in Appendix \ref{app:algorithm}. 
With this method, we can efficiently explore $\Theta$ to find parameters with low enough loss\footnote{In a practical setting, this means that the loss should be comparable to a parameter that has been found by known symmetries.} to be considered in $[\theta]$.
Given the points found by GENNI, we conjecture that the hyperplane they lie in contains further 'good' solutions, making it more suitable for visualisation than a randomly chosen subset.

To construct an $m$ dimensional hyperplane for a visualisation, we run GENNI until we find $m+1$ linearly independent vectors with low enough loss.
We then use Gram-Schmidt to produce an orthonormal basis $\{\tilde{\theta_i}\}_{i=1}^m$ for the hyperplane from the set of vectors $\{(\theta_i - \theta_0)\}_{i=1}^m$ centered at $\theta_0$.
Since we can only visualize a finite subset of the hyperplane, we define a grid of coefficients in $\mathbb{R}^{m}$ as $\{c^i\}_{i \in \mathcal{I}}$ where $\mathcal{I}$ is a finite index set for the grid\footnote{Generally, we use a grid which is defined by a grid bound $[a, b]$ and the number of grid points $n$ on that interval. 
The grid for a $m$ dimensional space is then be given by $\bigtimes_m \{a + \frac{i - 1}{n - 1} (b - a) \}_{1 \leq i \leq n}$, which is the Cartesian product of the grid on $[a, b]$.}.
To restrict our search to candidates that approximately belong to $[\theta]$, we choose some sufficiently small $\epsilon > 0$ and construct the set of vectors $[\theta]_{\epsilon; \mathcal{I}} = \{ \theta' | \theta' = \theta_0 + \sum_{i = 1}^m c^j_i \tilde{\theta_i} \in \Theta; J(\theta') < \epsilon; \forall j \in \mathcal{I} \}$, which we call the $\epsilon$-equivalent set.
We can also define the $\epsilon$-equivalent coefficient set as $C([\theta]_{\epsilon; \mathcal{I}}) = \{ c^j | \theta_0 + \sum_{i = 1}^m c^j_i \tilde{\theta_i} \in [\theta]_{\epsilon; \mathcal{I}} \}$.
Since $C([\theta]_{\epsilon; \mathcal{I}}) \subset \mathbb{R}^m$, it can be directly visualised if $m \leq 3$; otherwise, we have to resort to a dimensionality reduction tool, such as UMAP \cite{mcinnes2018umap}.
The runtime of creating such a set is $\mathcal{O}(N)$ where $N$ is the number of grid points. 
However, using standard griding techniques, the runtime will be exponential in the dimension $m$. 
For larger $m$ (approximately $ > 7$), we found that  visualisation was computationally infeasible while trying to be sensitive to the spacing of the grid and the value of $\epsilon$.

We note that there are other methods to locally explore the parameter space or speed-up the population of $[\theta]$ such as continue running SGD as a MCMC sampling method or continue along level sets of the loss $J(\theta)$. 
The disadvantage of these methods for visualization is that we are unable to restrict the subspace in which these solutions lie to be at most three dimensional; which we circumvent by constructing $C([\theta]_{\epsilon; \mathcal{I}})$.
We leave exploration of other methods for future work.  

\section{Experiments} \label{sec:experiments}

We consider a simple neural network architecture to compare our visualisations to theoretical expectations. 
Specifically, this is a feedforward network consisting of ReLU non-linearities with the model architecture given by $\phi(x, \theta) = \theta_2^T \relu(\theta_1 x)$ where $\theta_1, \theta_2 \in \mathbb{R}^2$ and $x \in \mathbb{R}$. We consider a more complex architecture in Appendix \ref{app:cnn}, where we replicate our results for the high dimensional LeNet \cite{lecun1998gradient}. 
We compare our visualisations to two theoretically well-understood symmetries for the feedward network: the permutation \cite{Sussman1992,pourzanjani2017improving} and the scaling symmetries \cite{Dinh2017}. 
The latter is due to the non-negative homogeneity property, where for $\alpha \in \R^+$, $\relu(\alpha x) = \alpha\relu(x)$. Therefore, we can scale the input weight to a node by a valid $\alpha$ and its output weight by $\alpha^{-1}$ without changing the function. This symmetry defines a topologically connected equivalence class.




\begin{figure}[!htbp]

\begin{subfigure}{0.5 \textwidth}
    \centering
    \includegraphics[width= 0.8\textwidth]{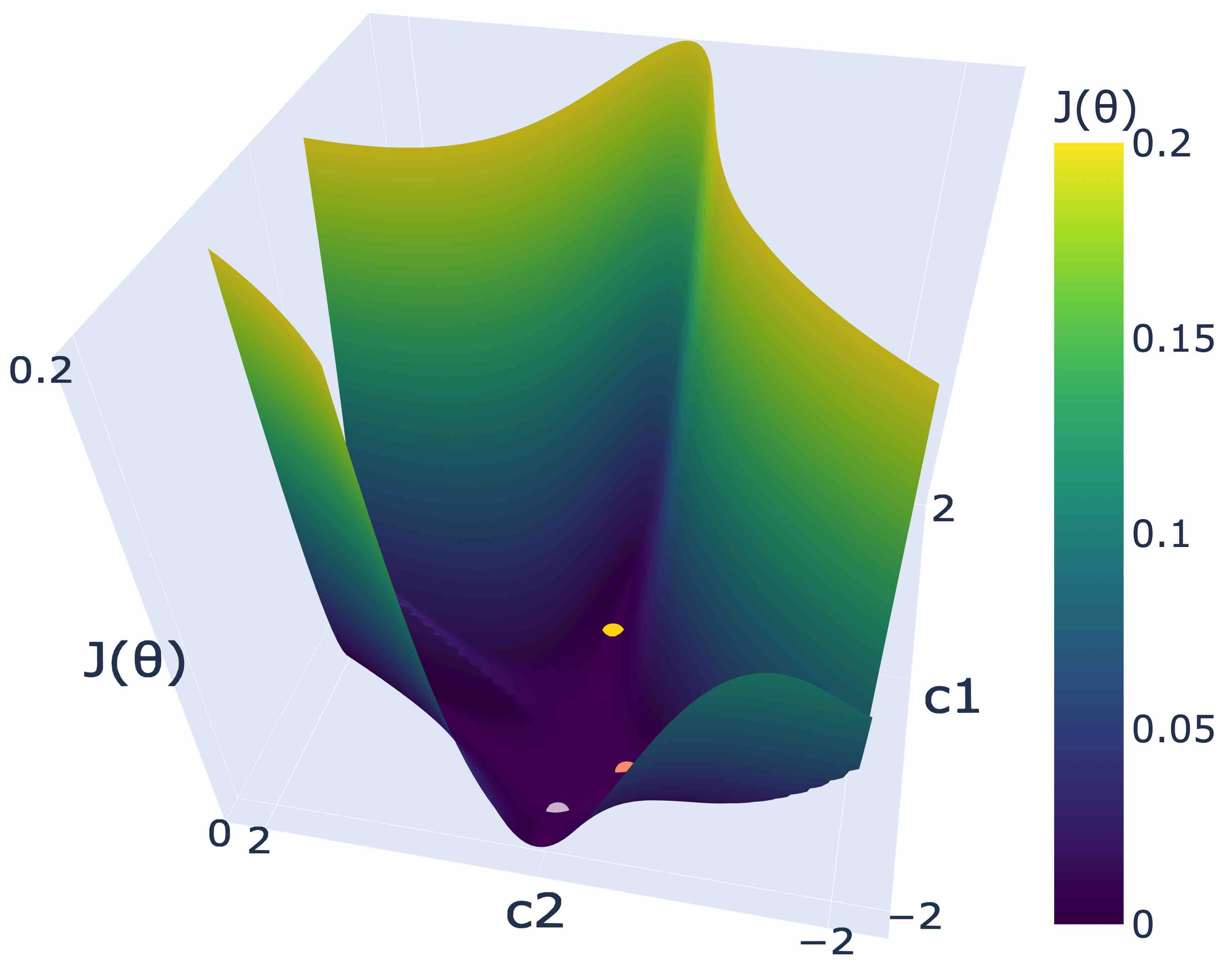}
\end{subfigure}%
\begin{subfigure}{0.5 \textwidth}
    \centering
    \includegraphics[scale=0.27]{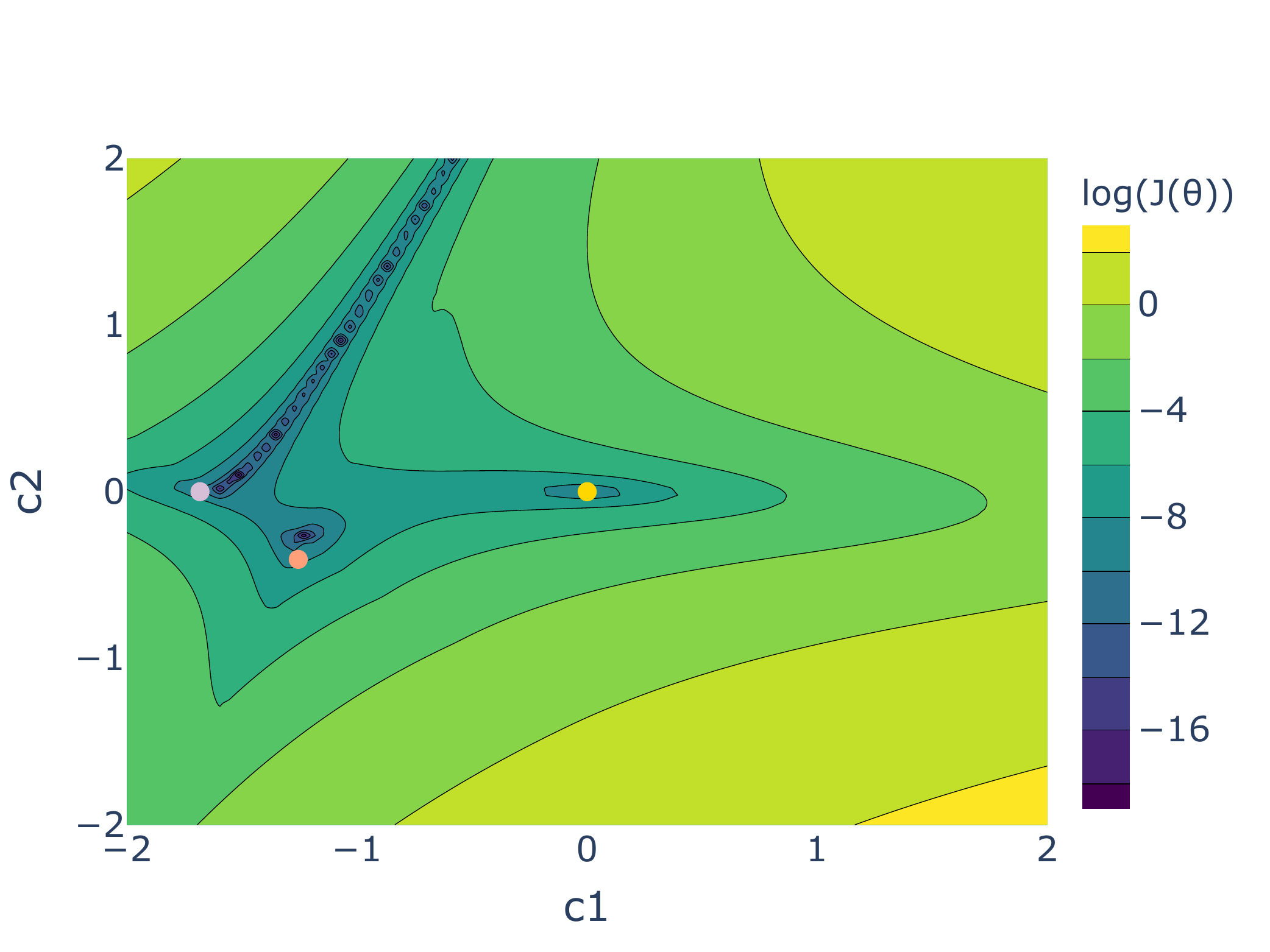}
\end{subfigure}%

\begin{subfigure}{0.5 \textwidth}
    \centering
    \label{fig:2d_surface_test}
    \subcaption{Surface Plot (linear scale)}
\end{subfigure}%
\begin{subfigure}{0.5 \textwidth}
    \centering
    \label{fig:2d_contour_test}
    \subcaption{Contour Plot (log scale)}
\end{subfigure}%

\caption{We fix a parameter vector $\theta$ and used our method as in Section \ref{sec:method}.
Then $c1$ and $c2$ represent the coordinates of the orthonormal basis vectors. We visualize the space via the auxiliary loss which allows us to visually infer the $\epsilon$-equivalent coefficient sets. We visualize $\theta_0, \theta_1$ and $\theta_2$ on the plot by denoting them by the red, purple and yellow dots.
We let the grid bound be $[-2, 2]$ with $100$ points.}
\label{fig:2d_test}
\end{figure}

\begin{figure}[!htbp]
\begin{subfigure}{0.5 \textwidth}
    \centering
    \includegraphics[width= 0.8\textwidth]{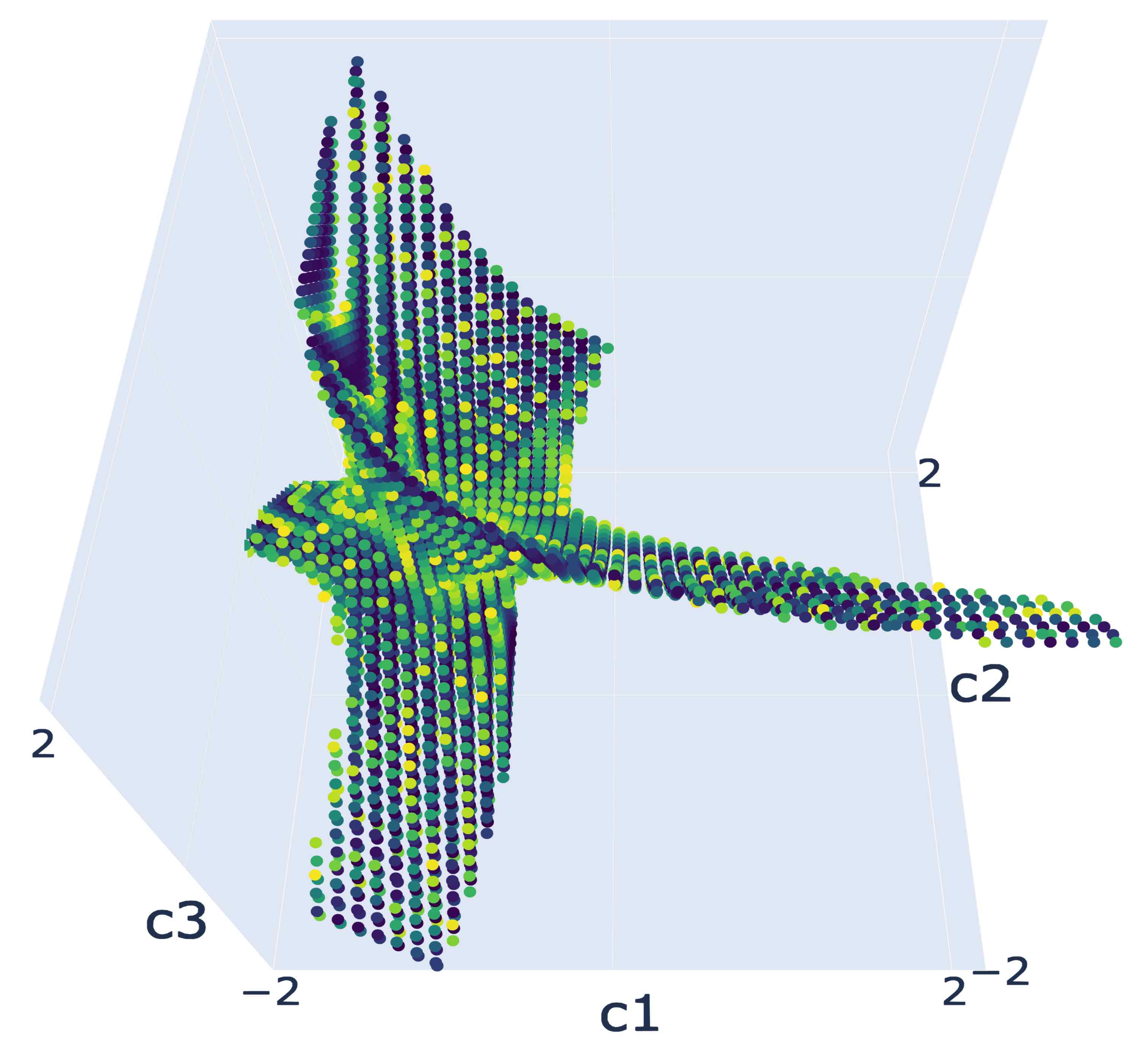}
    \label{fig:3d_angle1}
\end{subfigure}%
\begin{subfigure}{0.5 \textwidth}
    \centering
    \includegraphics[width= 1\textwidth]{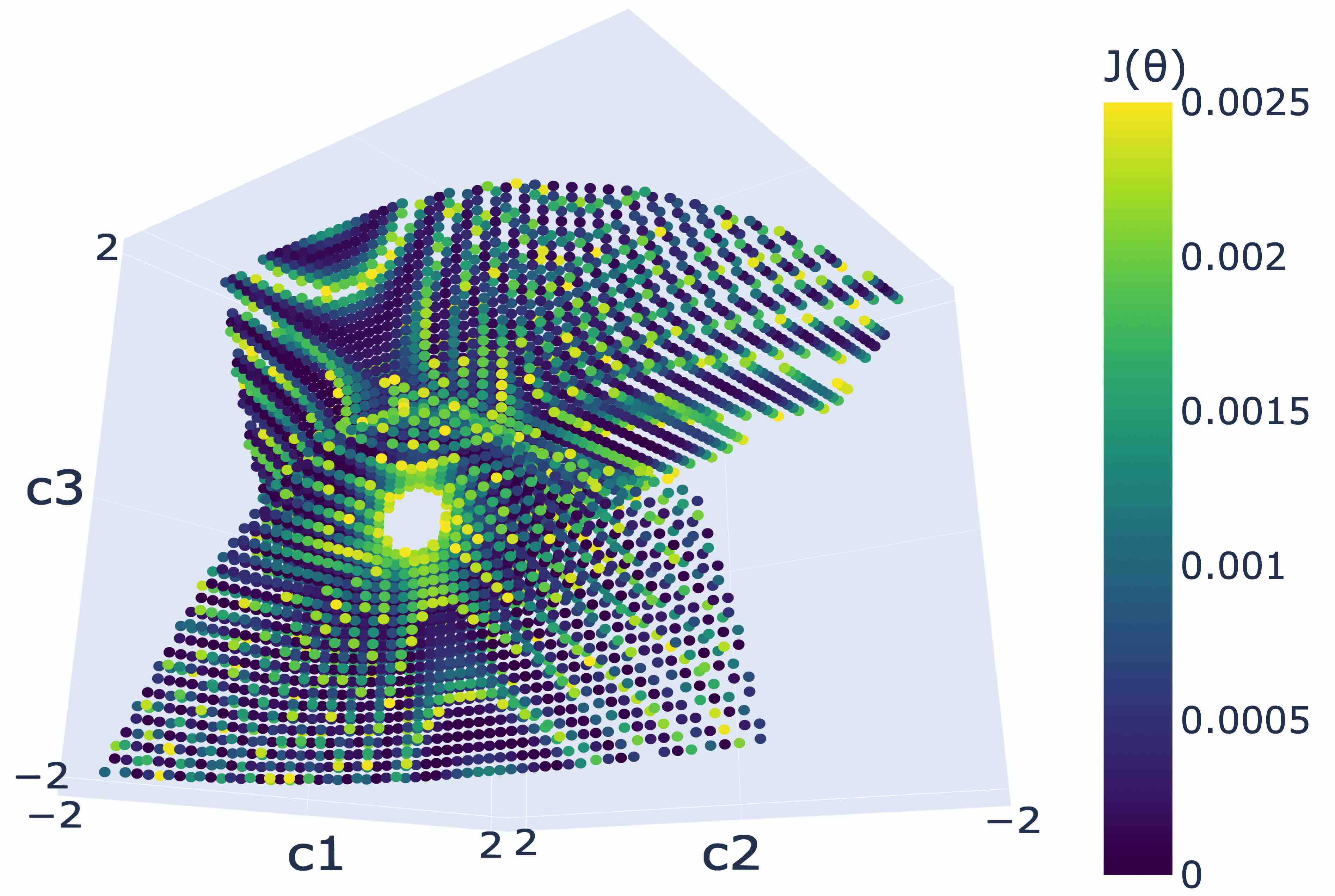}
    \label{fig:3d_angle2}
\end{subfigure}%

\caption{Similar to Figure \ref{fig:2d_test} but extended to the 3D case.
$c1, c2$ and $c3$ represent the coordinates of the orthonormal basis vectors. 
The color bar indicates the auxiliary loss for the parameter vector represented by $c1, c2, c3$ and also demonstrates our chosen value of $\epsilon = 0.0025$.
We plot the 3D plot from different angles to better visualize the shape of the $\epsilon$-equivalent coefficient set on the chosen subset. 
We let the grid bound be $[-2, 2]$ with $50$ points. 
}
\label{fig:3d_test}
\end{figure}

In Figure \ref{fig:2d_test}, we consider a 2D subspace of our parameter space and plot the auxiliary loss for each parameter vector. 
We can see that there is a connected region with low loss and the expected shape of $\alpha^{-1}$.
There also seem to be two distinct, yet similar regions of such shapes, which may be attributed to the permutation symmetries. 
However, in Figure \ref{fig:2d_test}, we do not observe the full shape with equally low loss.
This leads us to consider a three dimensional subspace in which we may have access to a more complete view of the equivalence class. 
In Figure \ref{fig:3d_test}, we clearly see two different intersecting surfaces in the shape of $\alpha^{-1}$.
Due to the permutation symmetries, such intersecting surfaces should be expected.
An interesting feature, which is not obviously explained by scaling and permutation symmetries is the hole in Figure \ref{fig:3d_test}(b); may be explained by other groups of symmetries, as discussed in \cite{petzka2020notes}. 

Lastly, we visualise higher-dimensional spaces via UMAP, as can be seen in Appendix \ref{app:umap_fcn}. Such visualisations do seem to be less interpretable however. 
For LeNet \cite{lecun1998gradient} we also observe connectedness albeit with a different geometry and the additional property of a non-empty interior, as can be seen in Appendix \ref{app:cnn}.


 \label{sec:results}

\section{Conclusions and Future Work} \label{sec:conclusions}
We propose a scaleable (in the size of the network) method for identifying and visualising equivalent classes of a given parameter vector; which can be used for studying the identifiability problem. We showcased its use on a simple network, where we observed some non-trivial properties of the equivalence class, in addition to the expected symmetries. 
In addition, GENNI can be used to proactively probe regions of interest to, among other applications, inform initialization techniques for deep learning tasks.
Instead of restricting the visualisations to only a hyperplane, GENNI could also be used as a Monte Carlo estimator to generate elements of an equivalence class.
However, this would require further and more principled dimensionality reduction techniques for visualisations.

Immediate future work is to develop an in-depth guide on how to use our visualisations to deduce the symmetries of a specific neural network architecture. We hope to extend such a guide to complicated neural networks, such as the one visualised and depicted in Appendix \ref{app:cnn}. Following this, we will explore how the quotient space, under the defined equivalence relation, looks like using GENNI. Further, we would like to explore extended applications, such as in understanding meta-learning \cite{Hospedales2020} and transfer learning \cite{pan2009survey}. Truly understanding meta-learning requires us to explore additional structure on a set of related models in the model space \cite{petangoda2020foliated}. With GENNI, the behaviour of meta-learning techniques can be better understood.





    

\clearpage


\begin{ack}
We would like to thank Berfin Simsek for presenting her work on weight space symmetries of neural networks \cite{brea2019weight} to us during the initial stages of this project.
\end{ack}
\bibliography{main}
\bibliographystyle{plain}

\appendix
\section{Algorithm} \label{app:algorithm}

\begin{algorithm}[H]
    \caption{GENNI: Calculate approximate equivalence class using SGD}
    \label{alg:approximate-equivalence-class}
    \begin{algorithmic}[1]
        \Procedure{Calculate-Approximate-Equivalence}{$\theta, C$}
            \State $[\theta]_{\text{Approx}} \gets \emptyset$
            \State Form auxiliary problem $J(\tilde{\theta}) = \frac{1}{\abs{C}}\sum_{x \in C}(\phi(x, \tilde{\theta}) - \phi(x, \theta))^2$
            \State{Generate $m$ starting points $(\theta_i)_{i=1}^m$}
            \For{$j = 1, \dots, m$}
                \State{$\tilde{\theta}_{i} \gets \text{SGD}(J, \theta_{i})$} \Comment{Run SGD on $J(\tilde{\theta})$ starting from $\theta_i$}
                \State $[\theta]_{\text{Approx}} \gets [\theta]_{\text{Approx}} \cup \{\tilde{\theta}_i\}$
            \EndFor
            \State \textbf{return} $[\theta]_{\text{Approx}}$
        \EndProcedure
    \end{algorithmic}
\end{algorithm}

\section{Hyperparameters} \label{app:hyperparameters}

\subsection{LeNet}
The architecture of the network used is as in \cite{lecun1998gradient}.
\begin{table}[H]
\caption{Hyperparameters for the LeNet experiments seen in Section \ref{app:cnn}}
\centering
\begin{tabular}{|l|l|}
\hline
\textbf{Hyperparameter} & \textbf{Value} \\ \hline
Batch size              & 256            \\ \hline
No. of data points       & 8192           \\ \hline
Learning rate           & 0.001          \\ \hline
Seed                    & 0              \\ \hline
Network input dim.      & $28 \times 28$        \\ \hline
Network output dim.     & $1 \times 1$          \\ \hline
Optimizer               & SGD            \\ \hline

\end{tabular}
\end{table}

\subsection{Fully Connected Network}
The fully connected network architecture has four trainable parameters $a, b, c, d \in \R$ and the form:
\begin{align*}
    \phi(x, (a, b, c, d)) &= c \mathrm{ReLU}(a x) + d \mathrm{ReLU}(b x).
\end{align*}
\begin{table}[H]
\caption{Hyperparameters for the simple experiments seen in Section \ref{sec:experiments}}
\centering
\begin{tabular}{|l|l|}
\hline
\textbf{Hyperparameter} & \textbf{Value} \\ \hline
Batch size              & 256            \\ \hline
No. of data points       & 16384           \\ \hline
Learning rate           & 0.015          \\ \hline
Seed                    & 10              \\ \hline
Network input dim.      & $1 \times 1$   \\ \hline
Network output dim.     & $1 \times 1$   \\ \hline
Optimizer               & SGD            \\ \hline
\end{tabular}
\end{table}

\section{Further Visualizations} \label{app:umap_fcn}

\begin{figure}[!htbp]

\begin{subfigure}{0.5 \textwidth}
    \centering
    \includegraphics[width=0.7\textwidth]{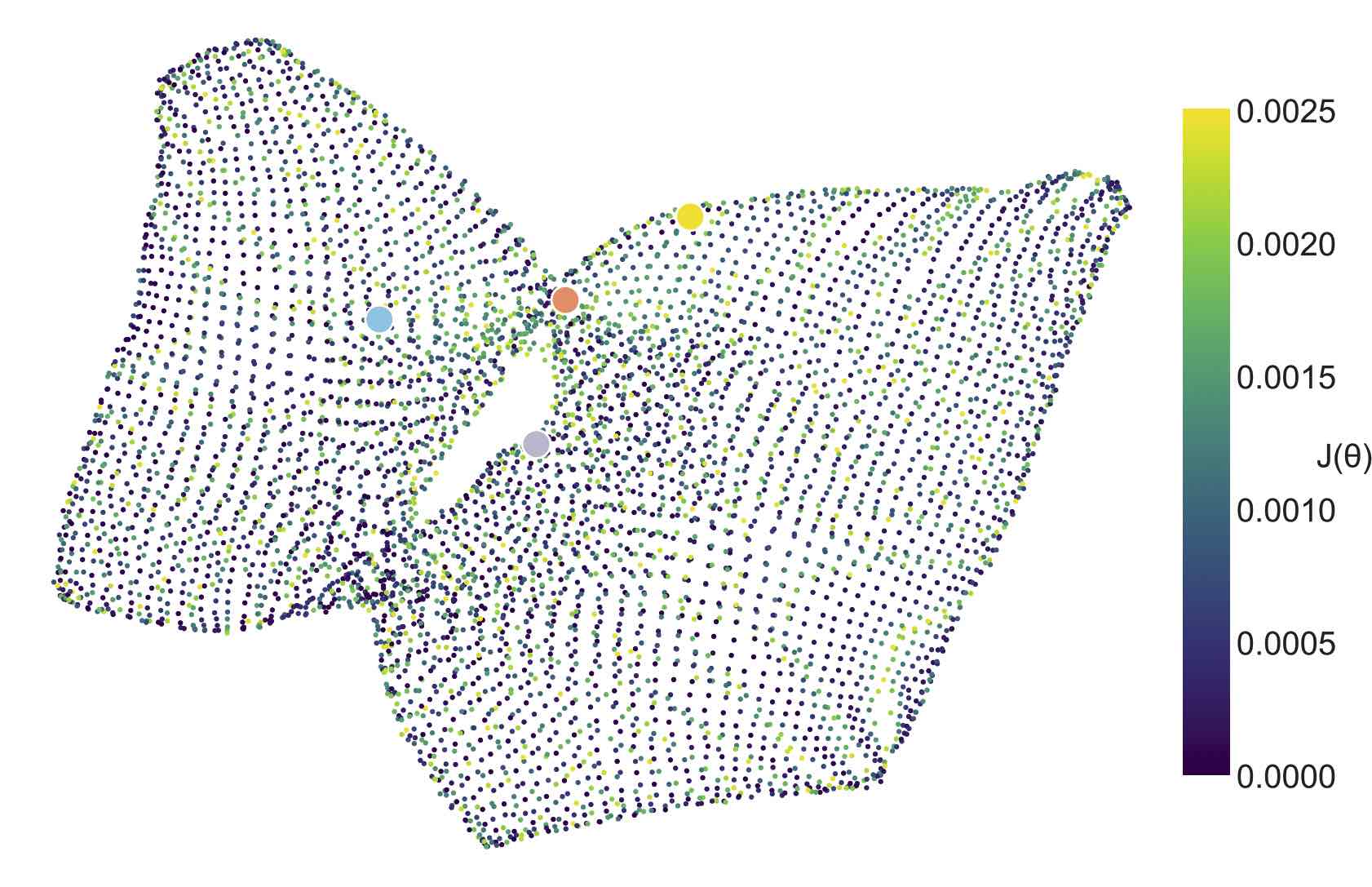}
\end{subfigure}%
\begin{subfigure}{0.5 \textwidth}
    \centering
    \includegraphics[width=0.7\textwidth]{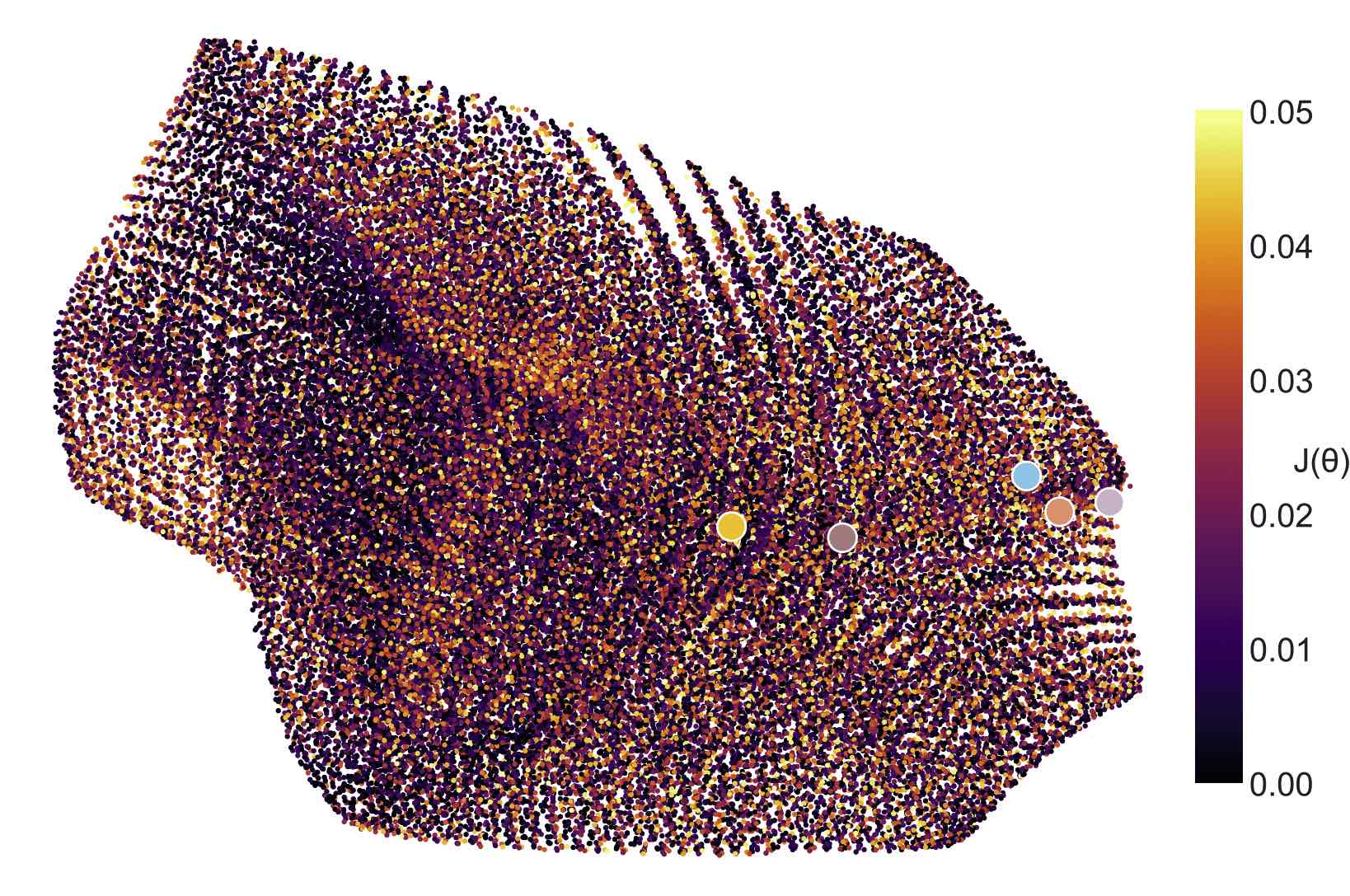}
\end{subfigure}%

\begin{subfigure}{0.5 \textwidth}
    \centering
    \subcaption{UMAP of a 3D subset.}
    \label{fig:UMAP_3D}
\end{subfigure}%
\begin{subfigure}{0.5 \textwidth}
    \centering
    \subcaption{UMAP of a 4D subset.}
    \label{fig:UMAP_4D}
\end{subfigure}%

\caption{In Figure \ref{fig:UMAP_3D} we use the same grid as in Figure \ref{fig:3d_test} but use UMAP to visualise $[\theta]_{\epsilon; \mathcal{I}}$. We also extend the same to 4D (we find an additional $\theta_4$ vector equivalent to $\theta$ via our gradient descent method). The color bars indicate the auxiliary loss for the parameter vectors and also demonstrate our chosen value of $\epsilon = 0.005$ for Figure \ref{fig:UMAP_3D} and $\epsilon = 0.1$ for Figure \ref{fig:UMAP_4D}.
We visualise $\theta_0$, $\theta_1, \theta_2, \theta_3$ and $\theta_4$ on the plot by denoting them by the blue, red, purple, yellow and brown dots. 
For the 3D figure we let the grid bound be $[-2, 2]$ with $50$ points. For the 4D case we let the bound be $[-1, 1]$ with $20$ points.} \label{fig:UMAP_test}
\end{figure}

In Figure \ref{fig:UMAP_3D}, we use UMAP to visualize the 3D space, and see that the shape of Figure \ref{fig:3d_test} is roughly maintained. 
Importantly, the region remains connected.
We also plot the equivalence class for a 4D subset in Figure \ref{fig:UMAP_4D}.
Here, the shape is generally less interpretable, but the region remains connected.
We do note that visualizations become much less explainable when using UMAP to move to higher dimensions, but that some topological properties, such as connectedness, may still be deduced. 



\section{Convolutional Neural Network Visualization} \label{app:cnn}

By considering a high-dimensional neural network, we can get a more complete picture of the benefits of our method. 
For that we consider the LeNet-5 \cite{lecun1998gradient} convolutional neural network architecture which has $60,000$ parameters. 
Due to the curse of dimensionality, it becomes even more difficult to simply grid the entire space to find equivalent parameters.
Hence, a more principled method for the selection of such subsets is necessary. This makes GENNI a good candidate. 

\begin{figure}[H]

\begin{subfigure}{0.5 \textwidth}
    \centering
    \includegraphics[width=0.7\textwidth]{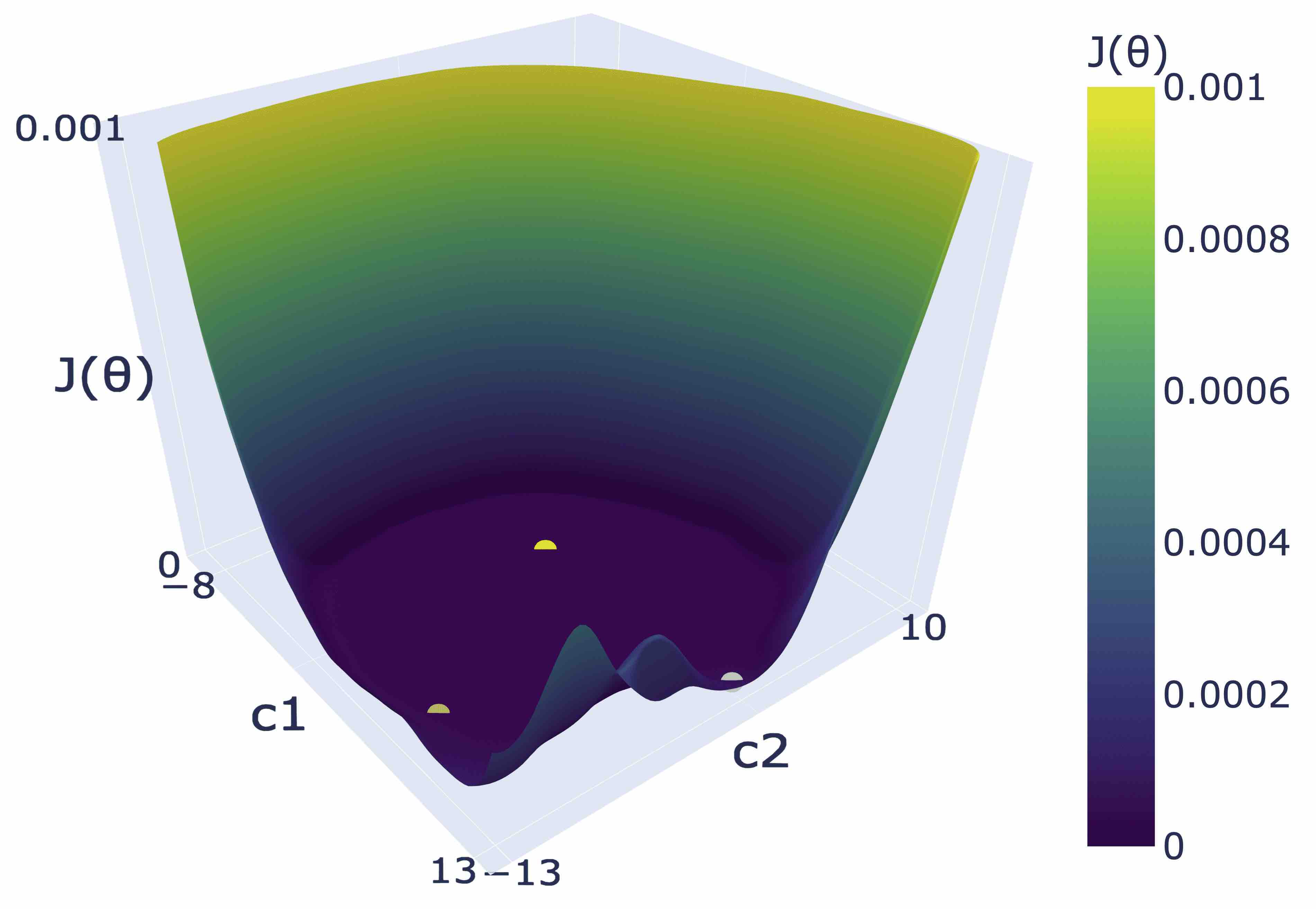}
\end{subfigure}%
\begin{subfigure}{0.5 \textwidth}
    \centering
    \includegraphics[scale=0.27]{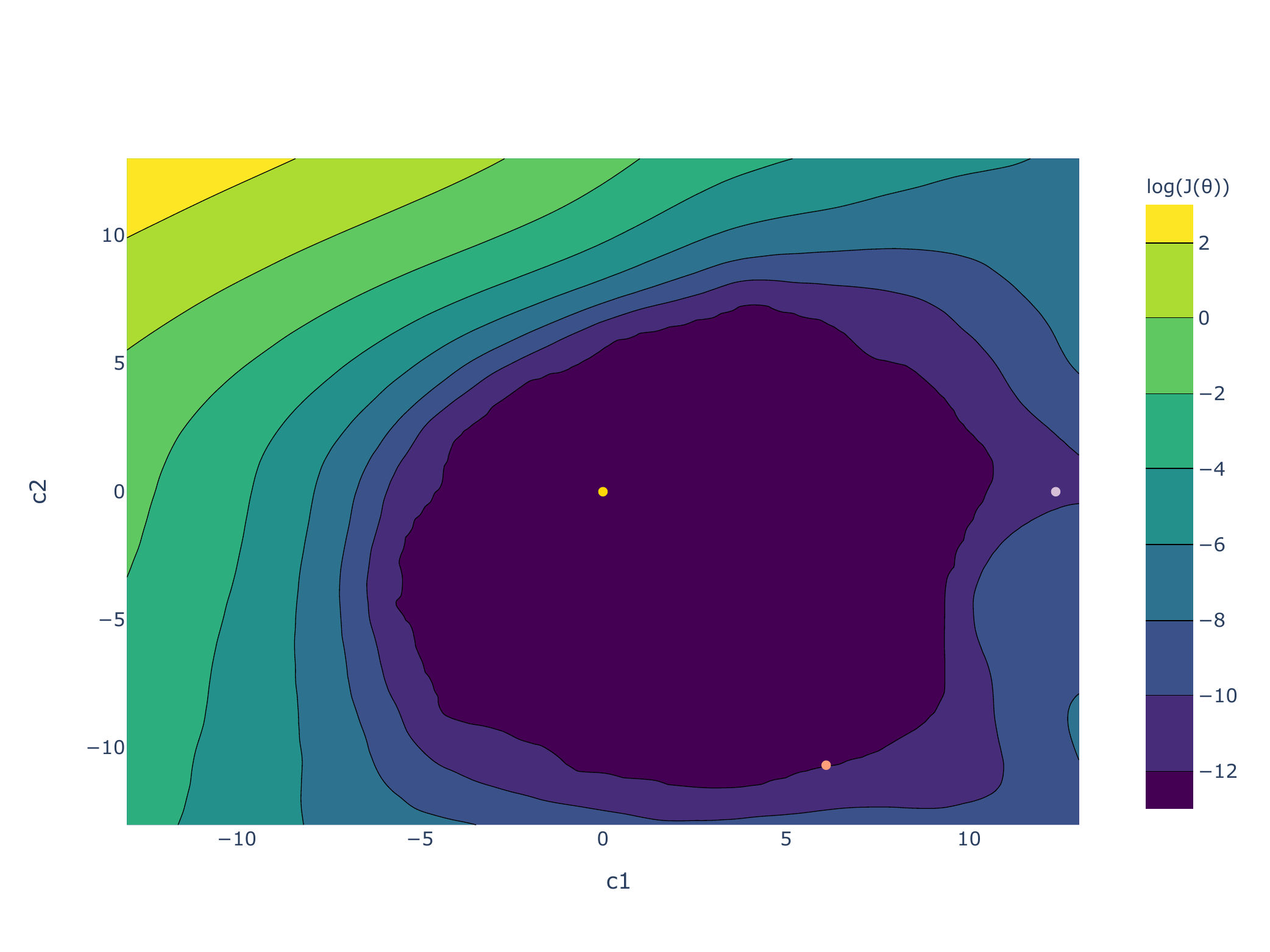}
\end{subfigure}%

\begin{subfigure}{0.5 \textwidth}
    \centering
    \label{fig:2d_surface_lenet}
    \subcaption{Surface Plot (linear scale)}
\end{subfigure}%
\begin{subfigure}{0.5 \textwidth}
    \centering
    \label{fig:2d_contour_lenet}
    \subcaption{Contour Plot (log scale)}
\end{subfigure}%

\caption{We fix a parameter vector $\theta$ and use our gradient descent method to find equivalent vectors $\theta_0, \theta_1$ and $\theta_2$. We grid the the hyperplane centered at $\theta_0$ as described in Section \ref{sec:method}. Then $c1$ and $c2$ represent the coordinates of the orthonormal basis vectors. We visualize the space via the auxiliary loss which allows us to visually infer the $\epsilon$-equivalent coefficient sets. We visualize $\theta_0, \theta_1$ and $\theta_2$ as the red, purple and yellow dots.
The grid bound was $[-13, 13]$ with $100$ points. }
\label{fig:2d_lenet}
\end{figure}

\begin{figure}[H]
\begin{subfigure}{0.5 \textwidth}
    \centering
    \includegraphics[width=0.75\textwidth]{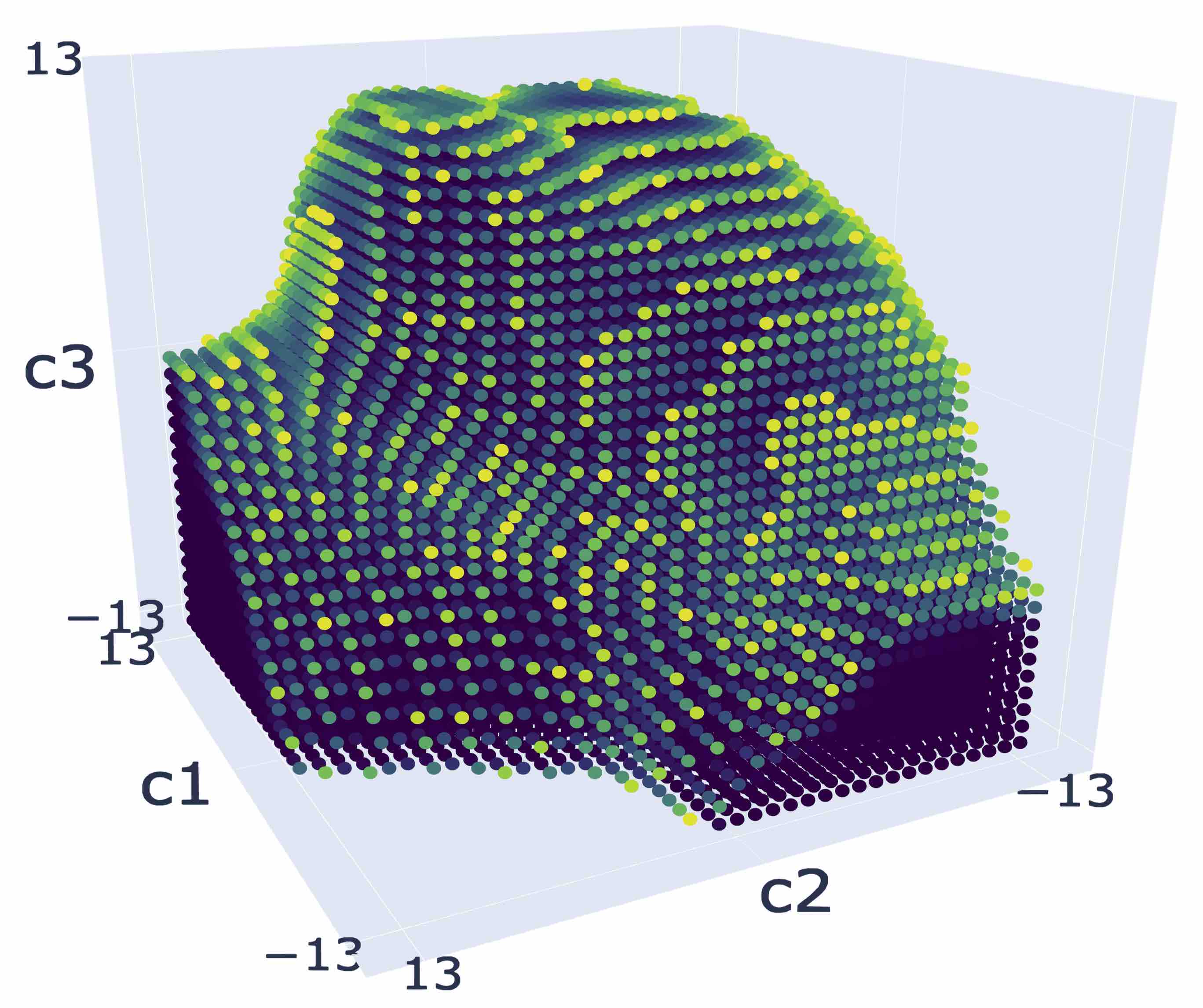}
    \label{fig:3d_angle1_lenet}
\end{subfigure}%
\begin{subfigure}{0.5 \textwidth}
    \centering
    \includegraphics[width=0.75\textwidth]{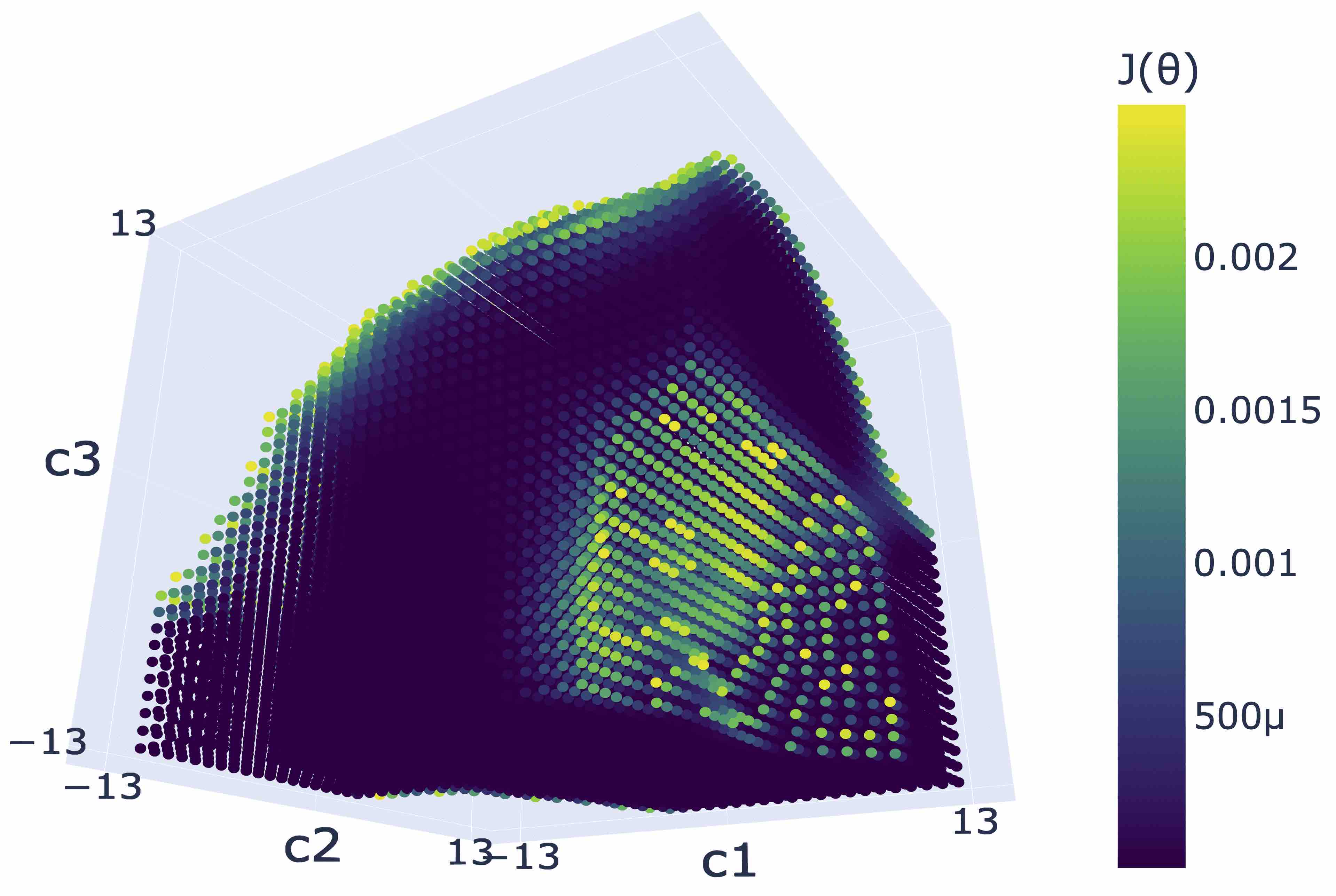}
    \label{fig:3d_angle2_lenet}
\end{subfigure}%

\caption{Similar to Figure \ref{fig:2d_lenet}, extended to the 3D case by using our gradient-descent method to find a $\theta_3$ which is equivalent to $\theta$.
$c1, c2$ and $c3$ represent the coordinates of the orthonormal basis vectors. 
The colorbar indicates the auxiliary loss for the parameter vector represented by $c1, c2, c3$ and also demonstrates our chosen value of $\epsilon = 0.0025$.
We plot the 3D plot from different angles to better visualize the shape of the $\epsilon$-equivalent coefficient sets. 
The grid bound was $[-13, 13]$ with $35$ points. 
}
\label{fig:3d_lenet}
\end{figure}

\begin{figure}[H]

\begin{subfigure}{0.5 \textwidth}
    \centering
    \includegraphics[width=0.75\textwidth]{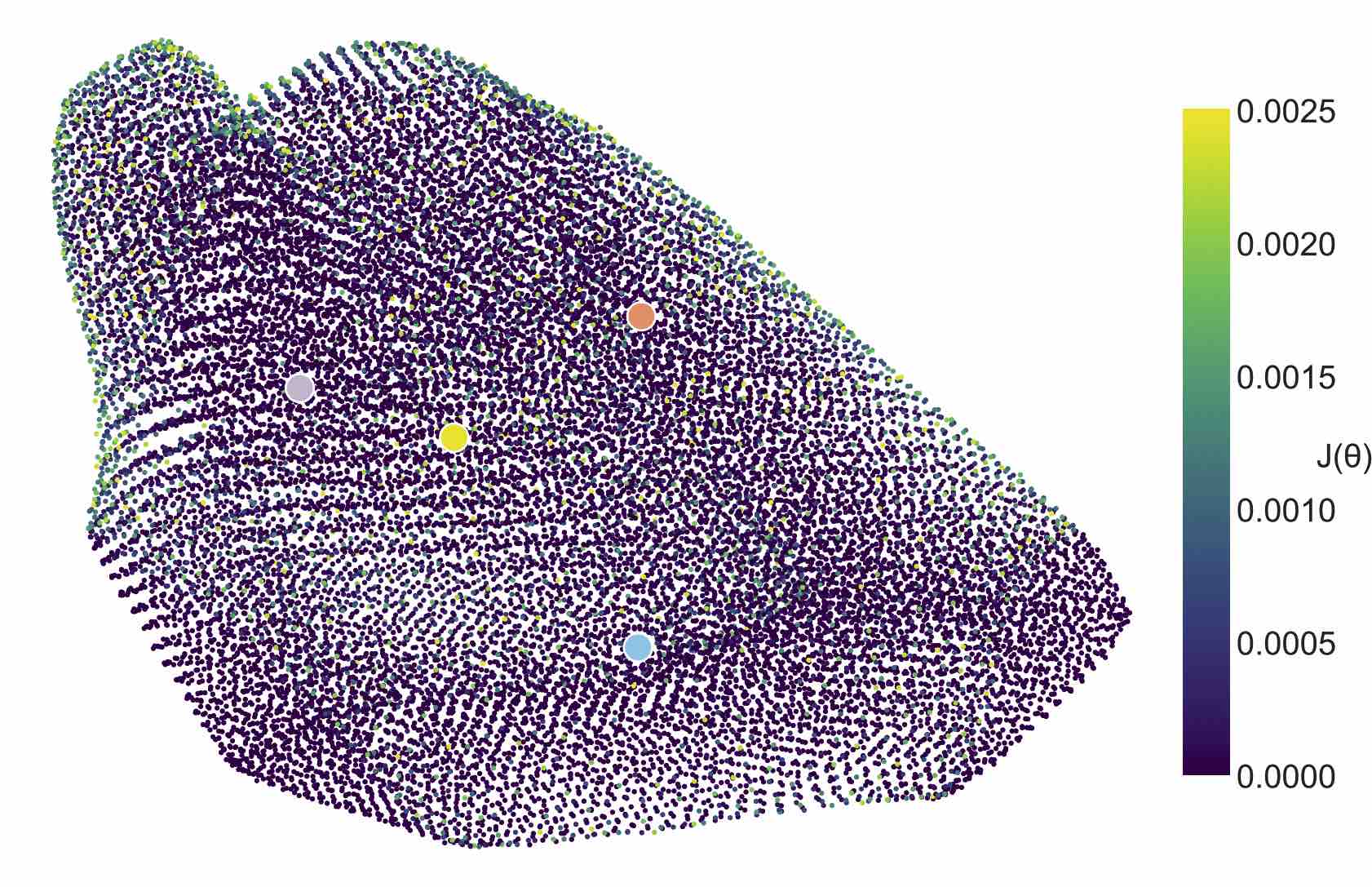}
\end{subfigure}%
\begin{subfigure}{0.5 \textwidth}
    \centering
    \includegraphics[width=0.75\textwidth]{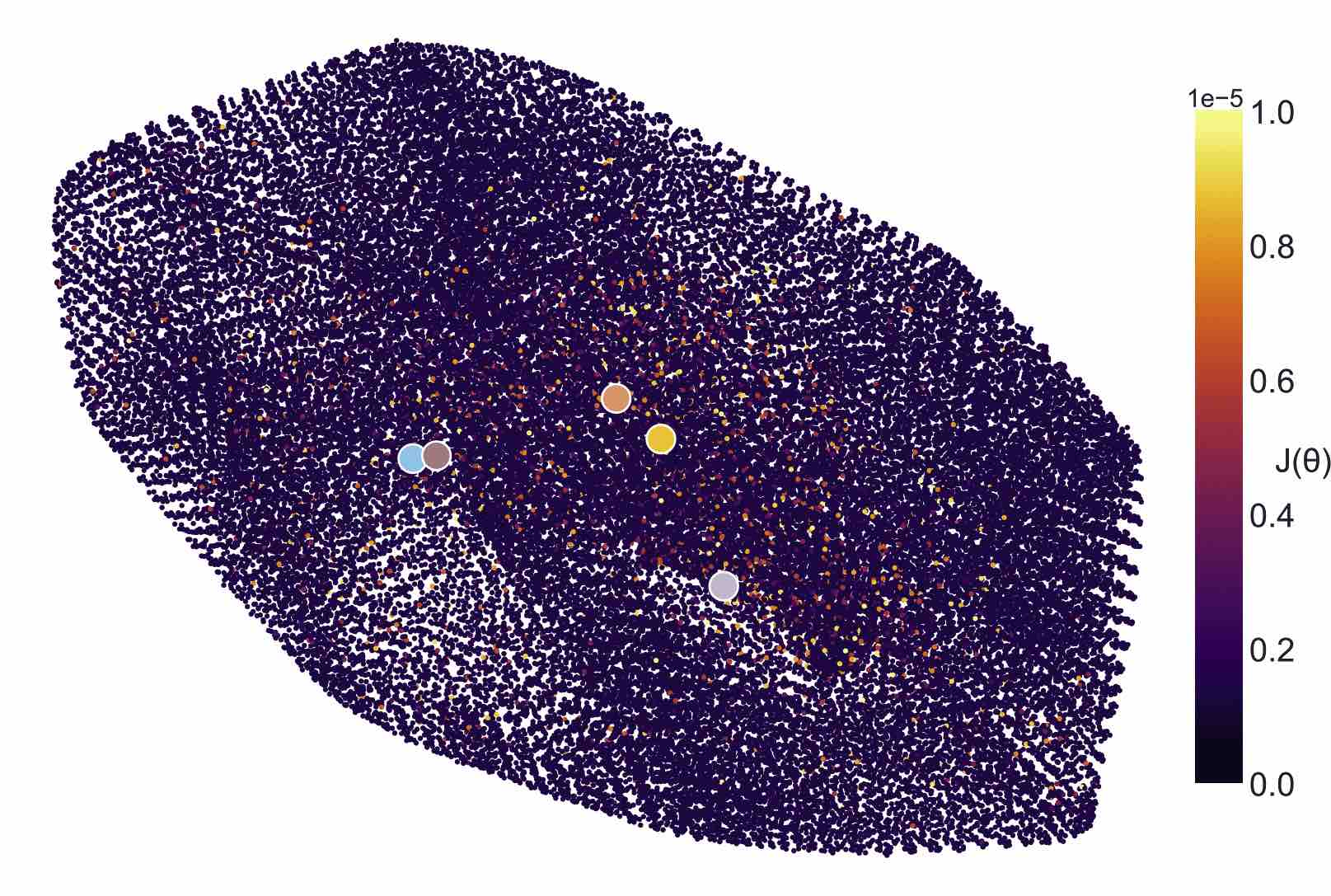}
\end{subfigure}%

\begin{subfigure}{0.5 \textwidth}
    \centering
    \subcaption{UMAP of a 3D subset.}
    \label{fig:UMAP_3D_lenet}
\end{subfigure}%
\begin{subfigure}{0.5 \textwidth}
    \centering
    \subcaption{UMAP of a 4D subset.}
    \label{fig:UMAP_4D_lenet}
\end{subfigure}%

\caption{In Figure \ref{fig:UMAP_3D_lenet} we use the same grid as in Figure \ref{fig:3d_lenet} but instead use UMAP to visualize the set of parameters $[\theta]_{\epsilon; \mathcal{I}}$, rather than the coefficients, in 2D. We also extend the same to 4D (we find an additional $\theta_4$ vector equivalent to $\theta$ via our gradient descent method) and again use UMAP to represent it in 2D. The colorbar indicates the auxiliary loss for the parameter vectors and also demonstrates our chosen value of $\epsilon = 0.0025$ for Figure \ref{fig:UMAP_3D_lenet} and $\epsilon = 10^{-5}$ for Figure \ref{fig:UMAP_4D_lenet}.
We visualize $\theta_0$, $\theta_1, \theta_2, \theta_3$ and $\theta_4$ as the blue, red, purple, yellow and brown dots. 
For the 3D figure, the grid bound was $[-13, 13]$ with $35$ points. For the 4D case, the grid bound was $[-1, 1]$ with $20$ points.} \label{fig:UMAP_lenet}
\end{figure}

We replicate the experiments in Section \ref{sec:results}. 
In Figure \ref{fig:2d_lenet} we can see that in contrast to the results for the feedforward network in Section \ref{sec:results}, the hyperplane has a large and wider connected region of low loss. 
We can see similar results in Figures \ref{fig:3d_lenet} and \ref{fig:UMAP_lenet}. 
The width of the equivalence class seems to be due to different symmetry groups acting on convolutional layers and due to the higher dimensionality of the parameter space.
However, we leave further investigations of symmetries to future work. 



\end{document}